\documentclass{article}

\usepackage{listings}
\usepackage{graphicx}
\usepackage{amssymb,amsmath,amsthm}
\usepackage{bbm}
\usepackage{xcolor}
\usepackage{xspace}
\usepackage{subcaption}

\newcommand{\lean}{{\tt Lean}\xspace}

\newcommand{\mathlib}{{\tt mathlib}\xspace}

\newcommand{\squeezedpara}[1]{\textbf{#1}}

\newcommand{\N}{\mathbb{N}}

\ifdefined\B
\renewcommand{\B}{\mathbb{B}}
\else
\newcommand{\B}{\mathbb{B}}
\fi

\usepackage{color}
\definecolor{keywordcolor}{rgb}{0.7, 0.1, 0.1}   %
\definecolor{commentcolor}{rgb}{0.4, 0.4, 0.4}   %
\definecolor{symbolcolor}{rgb}{0.0, 0.1, 0.6}    %
\definecolor{sortcolor}{rgb}{0.1, 0.5, 0.1}      %

\newcommand{\meas}{{\bf {\tt Meas}}}

\lstdefinelanguage{lean}{
  keywords = {def,let,return,noncomputable,in,theorem,lemma,end,begin},
  literate=
  {←}{{\ensuremath{\leftarrow}}}1
  {→}{{\ensuremath{\rightarrow}}}1
  {η}{{\ensuremath{\mathrm{\eta}}}}1
  {ν}{{\ensuremath{\mathrm{\nu}}}}1
  {γ}{{\ensuremath{\mathrm{\gamma}}}}1
  {ε}{{\ensuremath{\mathrm{\epsilon}}}}1
  {δ}{{\ensuremath{\mathrm{\delta}}}}1
  {θ}{{\ensuremath{\mathrm{\theta}}}}1
  {ℕ}{{\ensuremath{\mathbb{N}}}}1
  {ℍ}{{\ensuremath{\mathbb{H}}}}1
  {ℙ}{{\ensuremath{\mathbb{P}}}}1
  {ℝ}{{\ensuremath{\mathbb{R}}}}1
  {∧}{{\ensuremath{\wedge}}}1
  {×}{{\ensuremath{\times}}}1
  {∃}{{\color{symbolcolor}\ensuremath{\exists}}}1
  {μ}{{\ensuremath{\mathrm{\mu}}}}1
  {λ}{{\ensuremath{\mathrm{\lambda}}}}1
  {∩}{{\ensuremath{\cap}}}1
  {≥}{{\ensuremath{\geq}}}1
  {≤}{{\ensuremath{\leq}}}1
  {≠}{{\ensuremath{=}}}1
  {⊆}{{\ensuremath{\subseteq}}}1
  {∀}{{\color{symbolcolor}\ensuremath{\forall}}}1
  {∫}{{\ensuremath{\int}}}1,
  keywordstyle=[1]{\ttfamily\color{keywordcolor}},
  showstringspaces=false,
  keepspaces=true,
  basicstyle=\ttfamily\footnotesize,
  columns=[l]fullflexible,
  belowskip=-0.2 \baselineskip,
}

\lstdefinelanguage{leanin}{
  keywords = {def,let,return,noncomputable,in,theorem,lemma,end,begin},
  literate=
  {γ}{{\ensuremath{\mathrm{\gamma}}}}1
  {δ}{{\ensuremath{\mathrm{\delta}}}}1
  {ℕ}{{\ensuremath{\mathbb{N}}}}1
  {ℍ}{{\ensuremath{\mathbb{H}}}}1
  {ℙ}{{\ensuremath{\mathbb{P}}}}1
  {∧}{{\ensuremath{\wedge}}}1
  {×}{{\ensuremath{\times}}}1
  {μ}{{\ensuremath{\mathrm{\mu}}}}1
  {λ}{{\ensuremath{\mathrm{\lambda}}}}1
  {∩}{{\ensuremath{\cap}}}1
  {≤}{{\ensuremath{\leq}}}1
  {≥}{{\ensuremath{\geq}}}1
  {⊆}{{\ensuremath{\subseteq}}}1
  {∀}{{\color{symbolcolor}\ensuremath{\forall}}}1
  {∫}{{\ensuremath{\int}}}1,
  keywordstyle=[1]{\ttfamily\color{keywordcolor}},
  showstringspaces=false,
  keepspaces=true,
  basicstyle=\ttfamily,
  columns=[l]fullflexible,
}

\DeclareFixedFont{\ttb}{T1}{txtt}{bx}{n}{9} %
\DeclareFixedFont{\ttm}{T1}{txtt}{m}{n}{9}  %

\usepackage{color}
\definecolor{deepblue}{rgb}{0,0,0.5}
\definecolor{deepred}{rgb}{0.6,0,0}
\definecolor{deepgreen}{rgb}{0,0.5,0}

\newcommand\pythonstyle{\lstset{
language=Python,
basicstyle=\ttfamily\footnotesize,
otherkeywords={self},             %
keywordstyle=\ttb\color{deepblue},
emph={MyClass,__init__},          %
emphstyle=\ttb\color{deepred},    %
stringstyle=\color{deepgreen},
columns=[l]fullflexible,
showstringspaces=false            %
}}

\newcommand\pythonstylefile{\lstset{
language=Python,
basicstyle=\ttm,
otherkeywords={self},             %
keywordstyle=\ttb\color{deepblue},
emph={MyClass,__init__},          %
emphstyle=\ttb\color{deepred},    %
stringstyle=\color{deepgreen},
columns=[l]fullflexible,
showstringspaces=false            %
}}

\lstnewenvironment{python}[1][]
{
\pythonstyle
\lstset{#1}
}
{}

\lstnewenvironment{pythonin}[1][]
{
\pythonstylefile
\lstset{#1}
}
{}

\usepackage[nonatbib,preprint]{neurips_2020}
\makeatletter
\renewcommand{\@noticestring}{Preprint.}
\makeatother

\usepackage[utf8]{inputenc} %
\usepackage[T1]{fontenc}    %
\usepackage{hyperref}       %
\usepackage{url}            %
\usepackage{booktabs}       %
\usepackage{amsfonts}       %
\usepackage{nicefrac}       %
\usepackage{microtype}      %
\usepackage[numbers]{natbib}

\author{%
  Jean-Baptiste Tristan\\Oracle Labs
  \And
  Joseph Tassarotti\\Boston College
  \And
  Koundinya Vajjha\\University of Pittsburgh
  \AND
  Michael L. Wick\\Oracle Labs
  \And
  Anindya Banerjee\\IMDEA Software Institute 
}

\title{Verification of ML Systems via Reparameterization}

\begin{document}

\maketitle

\begin{abstract}
As machine learning is increasingly used in essential systems, it is
important to reduce or eliminate the incidence of serious bugs. A
growing body of research has developed machine learning algorithms
with formal guarantees about performance, robustness, or
fairness. Yet, the analysis of these algorithms is often complex, and
implementing such systems in practice introduces room for
error. Proof assistants can be used to formally verify machine learning systems by constructing
machine checked proofs of correctness that rule out such bugs.
However, reasoning about probabilistic claims inside of a proof assistant
remains challenging.
We show how a probabilistic program can be
automatically represented in a theorem prover using the concept of
\emph{reparameterization}, and how some of the tedious proofs of
measurability can be generated automatically from the probabilistic
program. To demonstrate that this approach is broad enough to handle
rather different types of machine learning systems, we verify both a
classic result from statistical learning theory (PAC-learnability of
decision stumps) and prove that the null model used in a Bayesian hypothesis
test satisfies a fairness criterion called demographic parity.
\end{abstract}

\section{Introduction}
\label{sec:introduction}
A machine learning application can fail for many reasons: maybe the
training data is insufficient, maybe there is a flaw in the design of
the learning algorithm, or maybe there is an error in the
implementation of the algorithm. Such errors can go unnoticed for long
periods of time. This is particularly worrisome for machine learning
applications that, for example, process loan requests or suggest
hiring recommendations.

Thorough testing is one way to help catch such errors. But testing ML
applications is challenging because of their random behavior. Many
iterations may be needed to encounter a bug or detect a statistical
irregularity in behavior.  And when it comes to adversarial or safety
critical scenarios, no amount of testing may be enough to make a
system trustworthy.  Moreover, while there has been much work on
developing algorithms that are provably robust or fair, bugs in
implementations of these algorithms may render these guarantees
meaningless.

One way to eliminate these kinds of errors is to \emph{formally
verify} a machine learning system with a machine checked proof of
correctness.
A formal proof is one in which every step and logical inference is checked.
The computer programs that help write and check such
proofs are called \emph{proof assistants}. Proof assistants provide a
language to express programs and mathematical proofs in some logic.
In recent years, it has become feasible
to use proof assistants to verify large, realistic software systems,
including compilers~\citep{leroy2009formal}, cryptographic primitives~\citep{fiat-crypto}, file systems~\citep{fscq-cacm}, and microkernels~\citep{sel4-cacm}.

In principle, proof assistants are expressive enough to represent the
mathematics underlying ML systems and check proofs of their
correctness. But in practice, formally verifying ML systems remains
challenging.  Prior work has begun to develop formal proofs of
correctness for machine learning software~\citep{selsam2017bugfree,
bagnall2019certifying}.
Although these early results are impressive, representing machine learning
programs and their correctness statements inside a proof assistant remains a
major challenge.  Proof assistants only have built-in support for representing
``pure'' mathematical functions. The approach taken in the aforementioned works
is to define, within the proof assistant, a small domain-specific programming
language that has commands for drawing random samples from distributions.

While this approach is feasible, it becomes challenging to use, particularly
when reasoning about samples from continuous probability distributions.
In particular, the way past work represents such programs in the proof assistant
is rather different from the way ML researchers usually think about things in pencil-and-paper proofs.

We propose using an alternative representation that more closely matches the familiar style of paper proofs.
Our approach is to
automatically \emph{reparameterize} parts of a probabilistic program
so that it can be written as a pure, non-randomized, functional
program operating over a pre-sampled list of random inputs.  This
simplifies reasoning about the resulting program and avoids the
foregoing difficulties. We have implemented an automatic translation
to perform this reparameterization, so that we can convert programs
from a probabilistic programming language based on Pyro \cite{Pyro}
into input to the theorem prover.

A simple example of our translation is shown
in \autoref{fig:intro_reparam}. On the left is a Python program
that samples from two distributions and returns a pair %
as result. On the right is our reparameterized version of this program in
the \lean proof assistant.  Background on \lean follows, but for now,
the important point is that in the translation, \texttt{majority\_fun}
takes as input an argument $\texttt{u}$ that represents a pair of
uniform $[0, 1]$ samples. Then instead of sampling, it uses the
inverse CDF transform to convert the input $\texttt{u}$ into the
distributions it needs. Finally, we define the distribution
represented by this program by taking the pushforward of the input
distribution through the function \texttt{majority\_fun}.

\begin{figure}

\begin{minipage}{0.35\textwidth}
\begin{python}
def majority():
  theta = uniform(0, 1) 
  X = bernoulli(theta)
  return (theta, X)
\end{python}
\subcaption[first caption.]{Python program}\label{fig:pyprog}
\end{minipage}
\quad
\begin{minipage}{0.5\textwidth}
\centering
\begin{lstlisting}[language=lean]
def majority_fun u :=
  let u1 := u.fst in
  let u2 := u.snd in
  let theta := gen_uniform(0,1,u1) in
  let X := gen_bernoulli(theta,u2) in
  (theta,X)

def majority :=
  push_forward majority_fun pair_uniform(0,1)

\end{lstlisting}
\subcaption[]{Reparameterization in \lean}\label{fig:reparam}
\end{minipage}
  \caption{Demonstration of our automatic reparameterization translation to represent programs in the proof assistant. Instead of drawing samples, {\tt majority\_fun} takes a pre-sampled pair {\tt u} of uniform [0, 1] variates. The expressions {\tt u.fst} and {\tt u.snd} are the first and second projections of the pair.}\label{fig:intro_reparam}
\end{figure}

To demonstrate our approach, we have verified two case studies. The
first is a proof that the class of decision stumps is
PAC-learnable~\citep{blumer1989learnability}, a classic introductory
result that appears in many textbooks on computational learning
theory. Formalizing this proof revealed errors and many omitted
details found in several expository accounts of this result.  Our
second case-study is a proof that a null model used in a Bayesian hypothesis test,
implemented in a probabilistic programming language, correctly satisfies
a fairness criterion called demographic parity. The source code
for our case studies and translator are publicly available.\footnote{\url{https://github.com/jtristan/FormalML}}

\section{A Short Introduction to Lean}
\label{sec:lean}
The \lean theorem prover can be viewed as both a functional
programming language (like Haskell) and a foundation for mathematics,
based on dependent type theory.  Dependent type theories are an
alternative to Zermelo-Frankel set theory where types are associated
with mathematical expressions, in the same way that types can be used
in programming languages, but with much stronger guarantees.  Before
we introduce the concept of dependent types, it is useful to consider
a simple example of mathematical formalization in \lean.

Using \lean as a programming language, we can define a function
\texttt{double} that takes a natural number as input and multiplies it by 2.
\begin{lstlisting}[language=lean]
def double(n: nat): nat := 2 * n\end{lstlisting}
This definition is similar to what one would find in any modern
functional programming language.  However, there is one significant
difference between programming in \lean and those languages: in order
to ensure that \lean is a consistent foundation for mathematics,
functions cannot have side effects (printing on the screen, reading a
file) and they must be proven to always terminate. Next, we can define
a predicate that formalizes the concept of an even number.
\begin{lstlisting}[language=lean]
def isEven(n: nat): Prop := exists k: nat, n = 2 * k\end{lstlisting}
This example clearly shows how \lean differs from a programming
language. The function we define does not return simple data like a
number or string, but instead a logical proposition that states that a
natural number $n$ is even if there exists a natural number $k$ such that $n = 2 *
k$.

\squeezedpara{Proofs with tactics:}
Finally, \lean lets us specify mathematical properties and prove
them. For example, the following states and proves a lemma called
\texttt{doubleIsEven} that says that the result of 
\texttt{double} is always even:
\begin{lstlisting}[language=lean]
lemma doubleIsEven: forall n: nat, isEven (double (n)) := 
begin 
  intros, unfold isEven, unfold double, existsi n, trivial, 
end\end{lstlisting}
The first line is the mathematical statement we wish to prove. What
follows the ``:='' and enclosed by the keywords ``begin'' and ``end''
is a set of commands, called tactics, that describes the proof in a
manner that \lean can check.  The programmer constructs this tactic
proof interactively: their IDE displays a list of current assumptions
and what remains to be proved.  This is represented by
a \emph{sequent}, which is a tuple of the form $\Gamma
\vdash \phi$, where $\Gamma$ is the list of hypotheses and variables (called the context)
and $\phi$ is a proposition (called the target). When the proof
starts, the sequent is $\emptyset \vdash \forall
n: \N, \texttt{isEven} (\texttt{double} (n))$. Executing the tactic
``intros'' transforms the sequent into $n: \N \vdash \texttt{isEven}
(\texttt{double}(n))$ where $n$ is now a fixed but arbitrary natural
number. Executing the tactic ``unfold'' applied to \texttt{isEven}
unfolds the definition of \texttt{isEven} to give the sequent
$n: \N \vdash \exists k: \N, \texttt{double}(n) = 2 * k$. Likewise, by
unfolding \texttt{double}, we obtain the sequent $n: \N
\vdash \exists k: \N, 2 * n = 2 * k$. Now we must
exhibit a choice for $k$ that satisfies the property, for which we can
use the tactic ``existi'' applied to $n$, which appears in the
context. This gives the sequent $n: \N \vdash 2 * n = 2 * n$ to which
we can apply the ``trivial'' tactic that ensures that a basic axiom
(namely, that equality is reflexive) has been reached.

\squeezedpara{\mathlib library:}
The \mathlib library is a large library of mathematical results
formalized in \lean~\citep{mathlib-cpp}.
In particular, it contains a formalization of
measure theory, based on \citet{HolzlH11}'s library from the Isabelle
theorem prover. Unfortunately, \mathlib does not a have a probability
theory library, and in order to formalize our results, we had to
develop one, as a special case of measure theory. This development
accounts for about 2,500 lines of \lean formalization.

\squeezedpara{Proof by reduction:}
An important feature of \lean that makes proofs easier is that the proof checker
can automatically execute or \emph{reduce} parts of programs. For example,
suppose at some point we need to show that the boolean expression
\texttt{(a and false) or (false and b)}
is always \texttt{false}. Instead of using lemmas for the basic rules
of boolean algebra, we can instruct \lean to case split on all
the possible values for \texttt{a} and \texttt{b}, evaluate the
expression, and then check that all cases reduce to \texttt{false}. Of
course, not all definitions can be executed or simplified this
way. For example, if we have a proof involving the Lesbesgue integral
from \mathlib, we cannot expect \lean to symbolically compute the
solution to an arbitrary definite integral. Definitions like the
integral are marked \texttt{noncomputable} in \lean, which means they
will not be executed.

\section{Denotational Semantics of Programs}
\label{sec:semantics}
With the probability theory library defined, we next need a
way to write down learning algorithms and probabilistic programs in \lean, so
that we can state and prove theorems about them, much as we did
with \texttt{double} above. However, by default, all functions in \lean have to
be purely functional (that is, they have to behave like mathematical functions,
with deterministic output and no side-effects).

\newcommand{\mbind}{\texttt{bind}}
\newcommand{\mret}{\texttt{ret}}
\newcommand{\mretw}{\textsf{ret}}
\newcommand{\mdo}[3]{\texttt{do}\; #1\leftarrow#2\, ;\;#3}

For that reason, past work on verifying machine learning algorithms
has used a \emph{denotational semantics} in which programs are
represented as distributions over the types of values they can return.
Given a type $X$, we write $\meas(X)$ for this type of probability
distributions on $X$. The first step is to define the primitive
distributions that our programs will need to sample from.  Next we
need a way to sequence together multiple steps of sampling and running
computations on sampled values. To do so, we can define a function
called \texttt{bind} of type $\meas(X) \rightarrow (X \rightarrow
\meas(Y)) \rightarrow \meas(Y)$.  That is, $\mathtt{bind}$ takes a
probability measure on $X$ and a function that transforms values from
$X$ into probability measures over $Y$, and returns a probability
measure on $Y$. Intuitively, we should read 
$\mathtt{bind(\mu, f)}$ as representing a program which first samples
from the distribution $\mu$ and then passes the result to $f$.  It is
common to use the notation $\mdo{x}{\mu}{g(x)}$ for $\mbind(\mu, g)$,
which helps reinforce the intuition that $\mbind$ samples from $\mu$
and then runs $g$.  We also define a function $\mathtt{ret}$ of type
$X \rightarrow \meas(X)$. It takes a value from $X$ and returns a
probability measure on $X$.

Functions $\mathtt{bind}$ and $\mathtt{ret}$ construct probability
measures, so their definitions say what probability they assign to an
event. If $A$ is an event we define them as:
\begin{align}
  \mbind(\mu,f)(A) & = \int_{x \in X} f(x)(A) d\mu \\
  \mret(x)(A) & = \chi_A(x)
\end{align}
While the definitions use standard mathematical notation, our
formalization uses the \lean definition of the Lebesgue integral, and
so on.  Here, $\mret(x)$ is the $\delta$-Dirac distribution at $x$. To
understand the definition of $\mbind$, consider the following
example. Let $\mu$ be a distribution over $X$ and consider the random
variables $U \sim \mu$ and $V \sim f(U)$. For example, $f$ could be
the function that for an input $l$ returns the distribution
$\mathcal{N}(l,0.1)$. What is the distribution of $V$? By the sum rule
of probability we have $\Pr(V=v) = \int_{u \in X} \Pr(V = v \mid U =
u) d\mu$. Therefore, $\Pr(V)(A) = \int_{u \in X} f(u)(A) d\mu$. Hence,
$\mbind(\mu,f)$ is simply computing the distribution that results from
applying $f$ while marginalizing over $\mu$.

\paragraph{Example.}
Consider the Python function from the introduction in
\autoref{fig:pyprog}.  The translation of this program into \lean as
we have so far described is:
\begin{lstlisting}[language=lean]
def majority1 :=
  do theta ← uniform(0, 1) ; 
  do X ← bernoulli(theta) ;
  ret (theta, X)\end{lstlisting}
The notation makes this look almost the same as the Python program
that we started with. However, in the \lean code, the ``function''
actually evaluates to a nested integral representing the distribution
that this procedure encodes. Similarly, in \lean, \texttt{uniform} and
\texttt{bernoulli} are not random number generators, but definitions
of those distributions in terms of their CDF. These \texttt{bind} and
\texttt{ret} operations are an example of a \emph{monad}. Monads are
commonly used in functional languages like Haskell to
represent programs that have side-effects. This probability monad was defined by
\citet{giry1982categorical}.

The Giry monad representation has some advantages.  As we have seen,
the denotation of a program has a structure that mimicks the original
source code. However, there are drawbacks when we try to reason about
programs expressed this way, particularly when we want to avoid using
axioms or making restrictions to discrete spaces. First, recall from
\autoref{sec:lean} that proof by reduction is blocked when we work
with noncomputable definitions like integrals. Because the Giry monad
includes an integral every time we use $\mathtt{bind}$, \lean cannot
reduce such programs very much at all. Second, in our experience, this
monadic semantics is unfamiliar to ML experts, and doesn't correspond
closely to the style used on paper proofs.  In the next section, we
describe our approach for alleviating these issues.

\section{Reparameterizing to Simplify Semantics}

\newcommand{\cc}[1]{\texttt{#1}} How can we find a denotational
semantics that would make it easier to reason formally about a
learning or randomized algorithm and avoid the issues with the Giry
monad described above? Recall that a classic result in probability
theory is that any distribution on the reals with the Borel sigma
algebra can be constructed as the pushforward of the uniform
distribution on $[0,1]$, which we write as $\mathcal{U}$. That is, for
any distribution $\mu$, there exists a (measurable) function $f$ such
that for any event $E$
\begin{align}\label{eqn:map_apply}
 \mu(E) & = \mathcal{L}[f^{-1}(E)]
\end{align}
Indeed, a more general version of this result is an important lemma
in proving the Kolmogorov Extension Theorem, which is used to
show the existence of many stochastic processes.

This fact hints at an alternate denotation for our programs: we could
represent a program as some pure function $f$, applied to
samples from $\mathcal{U}$. That is, we would take the pushforward
measure of $f$ applied to an appropriate input distribution. Because
$f$ will be pure, \lean will be able to evaluate it, enabling us to
use proof by reduction. Moreover, once we prove that $f$ is a
measurable function, we can avoid most measure-theoretic issues in the
proof.  Of course, the theorem above suggests that $f$ will exist
\emph{in principle}, but we still need a way to construct the function
$f$ in a useful form.

In order to help explain how we find a simple representation of $f$, we
first observe that such a function $f$ is a \emph{reparameterization}
of the original program.  A reparameterization is a transformation of
a probabilistic model that changes how a variable is sampled, often by
sampling an additional variable from an alternate distribution and
transforming the result.  For example, if $X$ is a random variable
with distribution $\mathcal{N}(3,4)$, we can reparameterize $X$ to
define it as $X = 2Z + 3$ where $Z$ is a draw from the standard normal
distribution. Reparameterization has many other applications in
ML. For example, it can improve the convergence of MCMC
algorithms~\citep{gorinovaMHarXiv}, and enable the use of
stochastic gradient descent in Variational Auto Encoders~\citep{KingmaW13}.

We now describe the steps to compute the reparameterization of
functions in our setting. We have implemented this translation for a
small probabilistic programming language implemented on top of
Pyro. However, to keep things self-contained, we will explain how the
translation works using simple Python programs, without dynamic
looping or recursion. Our $\cc{majority}$ function from
\autoref{fig:pyprog} will serve as a running example.

\squeezedpara{Transforming primitive distributions:}
First we extend the probability theory library in \lean to
include reparameterized definitions of all the primitive distributions
that our programs can sample from.  For example, \cc{gen\_uniform(a,
  b, u)} generates a uniform random variate on the interval $[a,b]$ by
scaling the input $\cc{u}$, which is assumed to be a uniform sample
from $[0, 1]$.  More generally, we can implement the inverse CDF
transform to convert a sample from the uniform distribution on $[0, 1]$ into the appropriate
distribution.

These primitive translations are added to a dictionary that tracks
their input and output types, which in particular records how many
uniform inputs they need.  As we translate a function, we replace
sampling from primitive distributions with these translations, and
record how many total uniform samples will be needed. Then, an
argument $\cc{u}$ is added that is a vector of all of the uniform
samples that will be needed. We write $\cc{u1}$, $\cc{u2}$, $\cc{u3}$
etc. for the components of this vector. These are defined using \lean's primitive
operations \cc{fst} and \cc{snd} for extracting the first and second element of a pair.

\squeezedpara{Slicing and Coupling:}
Next, for each return value of the function, we compute a
\emph{slice}, which is the subset of expressions in the function that
determine that return value.  For example, in the \cc{majority}
example from \autoref{fig:pyprog}, there are two return values,
\cc{theta} and \cc{X}. The slice for $\cc{theta}$ is
$\cc{gen\_uniform(0,1,u1)}$, and the slice for \cc{X} is
$\cc{gen\_bernoulli(gen\_uniform(0, 1, u1), u2)}$. Note that the
re-use of \cc{u1} in both the computation of \cc{theta} and \cc{X}
ensures that we properly capture the dependence between these two
random variables. In general, all dependencies between the variable
definitions are explicitly captured by which inputs are passed to
which parameters, so that the right joint distribution is obtained.
Essentially, the inputs
are being used to construct an appropriate \emph{coupling}~\citep{lindvall2002}
between the random variables encoded by the slices.
We generate \lean definitions for each slice,
and then call these to compute each return value in the function. Now,
we can perform standard compiler optimization transforms, such as
removing common sub-expressions, to simplify the function.

\squeezedpara{Nested functions:}
After we complete the translation of \cc{majority}, we can add an
entry to the dictionary of translations tracking its type.  When
translating a subsequent function \cc{g}, if a call to \cc{majority}
is encountered in \cc{g}, we can replace it with a call to the
translation and add additional uniform samples to the vector of inputs
to $\cc{g}$. Since the pre-translated version of \cc{majority} takes
no arguments as input, it is also possible to hoist the call to
\cc{majority} out of the body of \cc{g}, and instead pass in the
pre-sampled results of \cc{majority} as an argument to \cc{g}.
\autoref{fig:pyprog-parity} shows an example, where \cc{majority} is
called by \cc{demographic\_parity}. The transformed function takes as
input both a sample from \cc{majority} and a pair of uniform samples, which it uses
to generate $\cc{phi}$ and $\cc{Y}$.

\begin{figure}
\begin{minipage}{0.35\textwidth}
\begin{python}
def demographic_parity():
  [theta,X] = majority()
  phi = uniform(0.8 * theta,1)
  Y = bernoulli(phi)
  return (theta,X,phi,Y)
\end{python}
\subcaption[first caption.]{Python program}\label{fig:pyprog-parity}
\end{minipage}
\quad
\begin{minipage}{0.5\textwidth}
\begin{lstlisting}[language=lean]
def demographic_parity_fun u :=
  let theta := u.fst.fst in
  let X := u.fst.snd in
  let u3 := u.snd.fst in
  let u4 := u.snd.snd in
  let phi := gen_uniform(0.8 * theta,1,u3) in
  let Y := gen_bernoulli(phi,u4) in
  ((theta,X),(phi,Y))

def demographic_parity :=
  pushforward demographic_parity_fun
       (prod_measure majority pair_uniform(0,1))
\end{lstlisting}
\subcaption[first caption.]{Reparameterization in \lean}\label{fig:lean-parity}
\end{minipage}
\label{fig:bht}
\caption{\lean translation with nested functions for a Bayesian hypothesis testing example.}
\end{figure}

\squeezedpara{Automatic proof generation:}
In the course of generating the function $f$, we can also generate
proofs in \lean that the function $f$, and all the slices used to
compute the return values, are measurable. To do so, we first manually
wrote proofs of measurability for all primitive operations in the
language, as well as the transformer functions for the primitive
distributions, such as $\cc{gen\_uniform}$.  Then, using the fact that
the composition of two measurable functions is measurable, these
proofs for primitive operations %
are composed to produce a proof
that an entire function, such as $\cc{majority\_fun}$ is measurable.
This proof can then be re-used if $\cc{majority}$ is called in another
function.

\squeezedpara{Impact of reparameterization on proofs:}
Our motivation for defining a semantics based on reparamaterization
is to simplify formal proofs for learning and randomized
algorithms. By embedding programs as pure functions of random inputs,
we make it possible to use Equation~(\ref{eqn:map_apply}) to turn a
probabilistic statement on random variables into a statement on
events, getting the heavy machinery of measure-theoretic probability
theory out of the way. At that point, reasoning about the program
boils down to reasoning about the set of inputs to the program that
satisfy some properties, which is usually very basic, intuitive, and
allows us to apply reduction to the function to simplify the
reasoning. We will see an example of this in Section
\ref{sec:bayes_verif}.

\squeezedpara{Generality:}
A natural question is whether this kind of reparameterization
translation can be applied to more complex, general purpose
programs. As we have mentioned, results from measure theory suggest
that in principle this can be performed on a large class of programs.
But, for large programs with complicated control flow and looping, a
naive reparameterization translation may make the program harder to
understand.  However, even when reparameterization would be unnatural
to apply to the entire program, we believe that it can be useful to
apply to subcomponents. For example, consider a typical implementation
of stochastic gradient descent, which is usually structured as a
loop. Within each iteration of the loop, training examples are
randomly selected or ordered, and then gradients and updates are
computed. We can factor out and reparameterize the computation of gradients
to be pure functions. Then, use of the Giry monad would be limited to only
the remaining impure parts that glue together iterations of the loop.
This way, we would obtain the advantages of reparameterization for the bulk of the proof.

\section{Case Study 1: Decision Stumps}
For our first case study, we prove in \lean that the concept class of
decision stumps is PAC learnable.  We focus here on how this algorithm
is formulated in \lean. More complete details about the classic pencil-and-paper proof and our formalization are found in an earlier report~\citep{tassarotti:stump}.
Recall that a decision stump is a classifier that assigns binary
labels to real valued points based on whether they are above or below
some threshold value. Points above the threshold are labeled $0$, and
points $\leq$ are labeled 1.  We assume that there is some unknown
distribution $\mu$ of examples that the classifier will have to label,
and that the true labels of these examples are determined by some
threshold $t$, which is also unknown.  Training such a classifier is
straightforward: we take the maximum of all the training examples with
label $1$, and use that as our threshold.

To show that this decision class is PAC learnable, we must prove that
for all $\epsilon, \delta \in (0, 1)$, there is some number $n$ such
that with $n$ labeled training examples drawn independently from
$\mu$, this training algorithm gives a stump that achieves error rate
below $\epsilon$ with probability at least $1 -
\delta$. This is the one-dimensional version of the problem of learning an axis-aligned
rectangle, which is used as a motivating example and exercise in many
introductory texts on learning theory~\citep{kearns1994introduction,
  shalev2014understanding, mohri2018foundations}.

We first express the learning algorithm as a pure \lean
function \cc{choose}, which takes the training data as a vector of
examples, where each example is a pair consisting of the data point
and its label. The algorithm filters out the non-positive examples,
removes the remaining labels, and takes the maximum. The
parameter \cc{n} below tracks the number of examples in the
vector \cc{data}:
\begin{lstlisting}[language=lean]
def filter n data := vec_map (λ p, if p.snd then p.fst else 0) n data
def choose n data := max n (filter n data) 
\end{lstlisting}
The process of training on $n$ can then be described as taking the
$n$-ary product measure on the input distribution $\mu$, and then
first pushing-forward a function to assign the true labels, and then
pushing-forward the result with \cc{choose}.
\begin{lstlisting}[language=lean]
def denot :=
    let η := vec.prob_measure n μ  in 
    let ν := pushforward (label_sample target n) η in  
    pushforward (choose n) ν
\end{lstlisting}
Finally, to make claims about the error rate of this algorithm, we
define an event \cc{error\_set} which captures whether the label
assigned by a classifier \cc{h} differs from the true label provided
by the unknown \cc{target}. Finally, the \cc{error} rate of a
classifier is the probability of this set under the unknown
distribution $\mu$ of test examples:
\begin{lstlisting}[language=lean]
def error_set h := {x | label h x ≠ label target x}
def error h := μ (error_set h target)  
\end{lstlisting}

All of the randomization lies in the process of modeling the training
examples as if they have been sampled from some arbitrary training
distribution.  Because the training algorithm here is entirely
deterministic, we are able to write down this algorithm directly
without using the automated reparameterization we have
described.  Nevertheless, the experience of working with this formulation of the
algorithm convinced us of the benefits of working with a pure function
as much as possible, which led us to automating reparameteriztion to
handle examples such as the one we describe next.

\section{Case Study 2: Bayesian Hypothesis Tests}
\label{sec:bayes_verif}
In our next example, we prove a property of a null model used for a
Bayesian hypothesis test.  Recall that in Bayesian hypothesis testing,
we have models for how a data set may have been generated, along with
prior probabilities for those models. We use Bayes rule to update our
probabilities of the models and then select from among them~\citep{BF}.
In this case study, we consider the use of Bayesian hypothesis testing
to judge whether the output of a selection procedure satisfies a
fairness property called the four-fifths rule, which is used as a
criterion for disparate impact testing by the U.S. Equal Employment
Opportunity Commission \cite{guidelines}. In particular, the
four-fifths rule says that if the selection rate for a protected class
is less than 4/5 the rate for the majority group, then this can be
construed as evidence of violating legal standards for adverse impact.

In order to formulate a Bayesian hypothesis test for this criterion,
the first step is to write down a null model that is supposed to
satisfy the 4/5 test: that is, the selection rate is meant to
be \emph{at least} 4/5 the majority rate.
The \cc{demographic\_parity} example from \autoref{fig:pyprog-parity}
is this null model. The selection rate of the majority
class is \cc{theta}, and the selection rate of the minority class is \cc{phi}.  The model is constructed so that \cc{phi} is at least
$4/5 \cdot \cc{theta}$. Then, \cc{X} and \cc{Y} give the results of
the selection procedure on one member of the majority class and the
minority class, respectively. A more general version of this model
might draw different numbers of samples from the two classes. By
expressing this model in a probabilistic programming language such as
Pyro, and specifying a prior, we could then compare the posterior
probability, given some example data, to an alternative model.

Because the null model is supposed to represent a selection procedure that
satisfies the 4/5 test, it is important to prove that it in fact does. We
use \lean to prove this. We want to show that $0.8 \cdot \Pr[\cc{X} =
1] \leq \Pr[\cc{Y} = 1]$.  This follows from conditioning on \cc{theta}, so that
it suffices to show that for all $t \in [0,1]$, we have $ 0.8 \cdot \Pr[\cc{X} =
1 | \cc{theta} = t] \leq \Pr[\cc{Y} = 1 | \cc{theta} = t]$.\footnote{We use
conditional probability notation to explain the argument. More precisely, we are using a disintegrating measure~\citep{chang1997conditioning}.
Because we are conditioning
on the first projection of the product measure that we are pushing-forward, the existence
of this disintegration is a result of Fubini's theorem.}

For the \lean version of the demographic parity model, this is
formalized with the following events:
\begin{lstlisting}[language=lean]
def B(t: [0,1]) := {v | v.fst.fst = t}
def majority_selected := {v | v.fst.snd = 1}
def minority_selected := {v | v.snd.snd = 1}
\end{lstlisting}
We can divide the proof into two steps. Consider a Bernoulli random variable $\cc{Y'}$ with selection rate $4/5 \cdot \cc{theta}$, generated using the same $\cc{u4}$ that is used to generate $\cc{Y}$. First, we can show that $0.8 \cdot \Pr[\cc{X} = 1 | \cc{theta} = t] \leq \Pr[\cc{Y'} = 1 | \cc{theta} = t]$. This is simpler because having conditioned on $t$, the selection rate of \cc{Y'} is deterministic.

Second, we show that $\Pr[\cc{Y'} = 1 | \cc{theta} = t] \leq \Pr[\cc{Y} = 1 | \cc{theta} =t]$.
The key is that this part of the proof can be entirely reduced to proving a pure fact about {\tt demographic\_parity\_fun}. In particular, because probabilities are monotone with respect to subset ordering, we just have to show that the set of random inputs of the function {\tt
demographic\_parity\_fun} which cause $\cc{Y'} = 1$ is a subset of the inputs that cause ${\cc{Y}=1}$. This is captured by the following \lean statement, where we write {\tt f} as an abbreviation for
{\tt demographic\_parity\_fun}:
\begin{lstlisting}[language=lean]
∀ t≥0, set.prod {a: [0,1] × ℕ | a.fst = t}
           (set.prod [0,1] {a: [0,1] | generate_bernoulli(4/5 * t,a) = 1})
        ⊆ {a: ([0,1] × ℕ) × [0,1] × [0,1] | (f a).snd.snd = 1 ∧ (f a).fst.fst = t}
\end{lstlisting}

\section{Related Work}
\label{sec:related}
\squeezedpara{Measure-theory in proof assistants:}
There have been general formalizations of measure-theoretic
probability theory in a few proof assistants. \citet{hurd_thesis}
formalized basic measure theory in the HOL proof assistant, including
a proof of Caratheodory's extension theorem. \citeauthor{hurd_thesis}
uses a semantics that is closest to some of the ideas underlying our
reparameterization approach, in that he models randomized programs as
having access to an infinite ``tape'' of pre-sampled random
bits. \citet{HolzlH11} developed a more substantial library in the
Isabelle theorem prover.  \citet{AvigadHs14} use this library to
formalize a proof of the Central Limit Theorem. 

\squeezedpara{Machine-checked proofs for ML:}
We have already mentioned some more recent work that has formalized
machine learning results. \citet{selsam2017bugfree} use \lean to prove
the correctness of an optimization procedure for stochastic
computation graphs. They prove that the random gradients used in their
stochastic backpropagation implementation are unbiased. In their
proof, they add axioms to the system for various mathematical
facts. They argue that even if there are errors in these axioms that
could potentially lead to inconsistency, the process of constructing
formal proofs for the rest of the algorithm still helps eliminate
mistakes. \citet{bagnall2019certifying} give machine-checked proofs of
bounds on generalization errors. They use Hoeffding's inequality to
obtain bounds when the hypothesis space is finite or there is a
separate test-set and apply this result to bound the generalization
error of neural networks with quantized weights. Their proof is
restricted to discrete distributions and adds some results as axioms
(Pinsker's inequality and Gibbs'
inequality). \citet{bentkamp2019formal} formalize a result
by \citet{CohenSS16} which shows that deep convolutional arithmetic
circuits are more expressive than shallow ones. This result deals with
what is deterministically representable by these structures, so their
proof does not require probability theory.

\squeezedpara{Semantics of probabilistic programs:}
The representation of programs in \lean in the Giry monad is an
example of \emph{denotational semantics} \cite{giry1982categorical},
where the meaning of a program is given in terms of a mathematical
object (here, the distribution on the type of values it can
return). We have already alluded to some of the problems of using the
Giry monad and our reparameterization transform on programs with
complicated control flow or looping structure.  Defining denotational
semantics for probabilistic programs with arbitrary general recursion
and higher-order functions is challenging, and the subject of much
recent research~\cite{Staton,Tasson}.

\squeezedpara{ML for automated theorem proving:}
A related but distinct line of work applies machine learning
techniques to automatically construct formal proofs of theorems.  By
using a pre-existing corpus of formal proofs, supervised learning
algorithms can be trained to select hypotheses and construct proofs in
a formal system~\citep{holist, gamepad, holstep, SelsamB19}.
 
\section{Conclusion}
\label{sec:conclusion}
We present an approach to simplify reasoning about probabilistic
programs in proof assistants. By reparameterizing these programs to be
pure functions of pre-sampled randomized input, we can exploit more of
the native automation and support for reasoning about pure functions
found in many proof assistants. Our case studies show that our
approach can be applied to verify programs from a diverse range of
subfields of ML.

\bibliographystyle{abbrvnat}
\bibliography{ref}

\end{document}